\documentclass[letterpaper, 10 pt, conference]{ieeeconf}  %

\IEEEoverridecommandlockouts                              %

\usepackage{graphics} %
\usepackage{epsfig} %
\usepackage{times} %
\usepackage{amsmath} %
\usepackage{amssymb}  %
\usepackage{graphicx}
\usepackage{booktabs}
\usepackage{caption}
\usepackage{subcaption}
\usepackage{url}

\usepackage{letltxmacro}
\usepackage{colortbl}
\usepackage{multirow}
\usepackage{soul}
\usepackage{makecell}

\usepackage[many]{tcolorbox}
\usepackage{color,xcolor}
\definecolor{citecolor}{HTML}{0071bc}
\definecolor{mydarkblue}{rgb}{0,0.08,1}
\definecolor{mydarkgreen}{rgb}{0.02,0.6,0.02}
\definecolor{mydarkred}{rgb}{0.8,0.02,0.02}
\definecolor{mydarkorange}{rgb}{0.40,0.2,0.02}
\definecolor{mypurple}{RGB}{111,0,255}
\definecolor{myred}{rgb}{1.0,0.0,0.0}
\definecolor{mygold}{rgb}{0.75,0.6,0.12}
\definecolor{mydarkgray}{rgb}{0.66, 0.66, 0.66}

\definecolor{darkblue}{rgb}{0,0.08,1}
\definecolor{darkgreen}{rgb}{0.02,0.6,0.02}
\definecolor{darkred}{rgb}{0.8,0.02,0.02}
\definecolor{darkorange}{rgb}{0.40,0.2,0.02}
\definecolor{darkpurple}{RGB}{111,0,255}

\newcommand{\todocite}[1]{{\color{blue}{[citation needed]}}}

\definecolor{purple}{RGB}{160, 32, 240}
\definecolor{washblue}{RGB}{186, 224, 228}
\definecolor{sky}{RGB}{128, 128, 128}
\definecolor{seagreen}{RGB}{60, 179, 113}
\definecolor{building}{RGB}{128, 0, 0}
\definecolor{road}{RGB}{128, 64, 128}
\definecolor{sidewalk}{RGB}{0, 0, 192}
\definecolor{fence}{RGB}{64, 64, 128}
\definecolor{vegetation}{RGB}{128, 128, 0}
\definecolor{car}{RGB}{64, 0, 128}
\definecolor{sign}{RGB}{192, 128, 128}
\definecolor{pedestrian}{RGB}{64, 64, 0}
\definecolor{cyclist}{RGB}{0, 128, 192}

\def\be {\begin{equation}}
\def\ee {\end{equation}}
\def\beas {\begin{eqnarray*}}
\def\eeas {\end{eqnarray*}}
\def\bea {\begin{eqnarray}}
\def\eea {\end{eqnarray}}
\def\bes {\begin{equation*}}
\def\ees {\end{equation*}}

\newcommand{\bbR}{{\mathbb{R}}}

\newcommand{\bI}{\mathbf{I}}

\newcommand{\bs}{\mathbf{s}}

\newcommand{\bz}{\mathbf{z}}

\newcommand{\bc}{\mathbf{c}}

\usepackage{amssymb}%
\usepackage{pifont}%
\newcommand{\cmark}{\color{mydarkgreen}\ding{51}}%
\newcommand{\xmark}{\color{red}\ding{55}}%

\makeatletter
\usepackage{xspace}
\def\@onedot{\ifx\@let@token.\else.\null\fi\xspace}
\DeclareRobustCommand\onedot{\futurelet\@let@token\@onedot}

\def\eg{\emph{e.g}\onedot} 
\def\ie{\emph{i.e}\onedot}

\makeatother

\usepackage{color,xcolor}
\usepackage[normalem]{ulem}
\usepackage{ifthen}

\newcommand{\method}{{LidarDM}}

\definecolor{gold}{rgb}{1.0, 0.87, 0.0}
\definecolor{silver}{rgb}{0.75, 0.75, 0.75}
\definecolor{bronze}{rgb}{0.8, 0.5, 0.2}

\newboolean{showremoved}
\setboolean{showremoved}{false} %

\newcommand{\removed}[1]{%
  \ifthenelse{\boolean{showremoved}}{\textcolor{red}{\sout{#1}}}{}%
}

\title{\LARGE \bf
LidarDM: Generative LiDAR Simulation in a Generated World
}

\author{Vlas Zyrianov$^{1*}$ \and %
Henry Che$^{1*}$ \and %
Zhijian Liu$^{2}$ \and  %
Shenlong Wang$^{1}$%
\thanks{This project is supported by NSF Awards \#2331878, \#2340254, and \#2312102 and Intel, IBM, Amazon, NCSA, and Meta. 
}%
\thanks{* Both authors contribute equally to the research.}
\thanks{
$^{1}$ V. Zyrianov, H. Che, S. Wang are with UIUC. 
        Email: \{\tt\small vlasz2, hungdc2, shenlong\}@illinois.edu}%
\thanks{$^{2}$ Z. Liu is with NVIDIA. Email: 
        {\tt\small zhijianl@nvidia.com}}%
}

\begin{document}

\maketitle
\thispagestyle{empty}
\pagestyle{empty}

\begin{abstract}
We present LidarDM, a novel LiDAR generative model capable of producing \textit{realistic}, \textit{layout-aware}, \textit{physically plausible}, and \textit{temporally coherent} LiDAR videos. LidarDM stands out with two unprecedented capabilities in LiDAR generative modeling: (i) LiDAR generation guided by driving scenarios, offering significant potential for autonomous driving simulations, and (ii) 4D LiDAR point cloud generation, enabling the creation of realistic and temporally coherent sequences. At the heart of our model is a novel integrated 4D world generation framework. Specifically, we employ latent diffusion models to generate the 3D scene, combine it with dynamic actors to form the underlying 4D world, and subsequently produce realistic sensory observations within this virtual environment. Our experiments indicate that our approach outperforms competing algorithms in realism, temporal coherency, and layout consistency. We additionally show that LidarDM can be used as a generative world model simulator for training and testing perception models. We release our source code and checkpoints at \url{https://github.com/vzyrianov/LidarDM}

\end{abstract}

\section{INTRODUCTION}
\label{sec:intro}

Generative models are notable in areas such as image and video generation~\cite{stylegan, stablediffusion, dalle2, imagen}, 3D generation~\cite{get3d, dreamfusion, li2023instant3d}, compression~\cite{balle2016end, huang2020octsqueeze}, and editing~\cite{sdeedit, srgan}. More recently, they have shown great promise in robotics by generating realistic scenarios and sensory data for training and validating embodied agents, significantly reducing the cost of real-world experiments~\cite{mapprior, lidarsim, geosim, scenegen, trafficgen, drivegan, unisimberkeley, dreamitate, swerdlow2023street, shen2023sim, videoldm, yu2023scaling}. 

While advancements in conditional image and video generation~\cite{kim2023neuralfield, gaia, luo2023latent, esser2023structure} have been remarkable, the specific task of generatively creating scenario-specific, realistic LiDAR point cloud sequences for autonomous driving application remains under-explored. 
Current LiDAR generation methods fall into two broad categories, each of which suffers from specific challenges. (i) \textit{LiDAR generative modeling methods}~\cite{lidargen, ultralidar, caccia, zhang2023learning} are currently limited to single-frame generation and do not provide the means for semantic controllability and temporal consistency. (ii) \textit{LiDAR resimulation}~\cite{carla, tallavajhula2018off, fang2020augmented, lidarsim, unisim, lidarnerf} relies on user-created or real-world collected assets, this induces a high cost, restricts diversity, and limits broader applicability.

\begin{figure*}[t] \centering  %
\vspace{0.5em}
\includegraphics[width=0.8\textwidth]{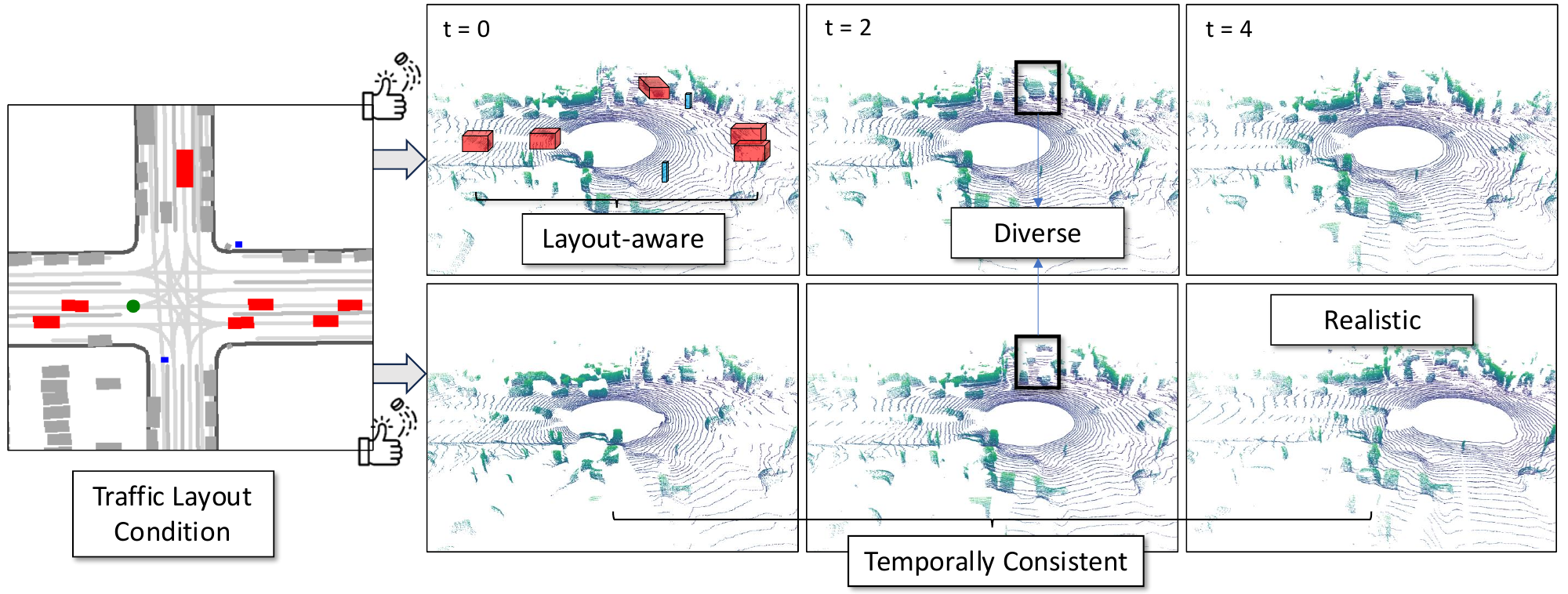}
 \caption{
 LiDAR sequences generated by LidarDM are realistic, layout-conditioned, physically plausible, and temporally coherent.
 We show 3 frames of 2 different videos generated from the same layout. In the layout visualization, the red rectangles are dynamic vehicles, gray rectangles are static vehicles, blue dots are pedestrians, the green dot is the ego vehicle sensor, and the gray curves are road boundaries and lane markings.). 
 }
 \label{fig:teaser}
 \vspace{-2em}
 \end{figure*}

To address these challenges, we propose LidarDM (Lidar Diffusion Model), which creates \textit{realistic}, \textit{layout-aware}, \textit{physically plausible}, and \textit{temporally coherent} LiDAR videos. LidarDM enables two previously unexplored capabilities : (i) controllable LiDAR synthesis guided by driving scenarios, which holds immense potential for simulation in autonomous driving, and (ii) 4D LiDAR synthesis for generating realistic and temporally coherent sequences of labeled LiDAR point clouds. Our key insight lies in first generating and composing the underlying 4D world and then creating realistic sensory observations within this virtual environment. We develop a novel approach for large-scale 3D scene generation based on the latent diffusion model and integrate existing 3D object generators for dynamic actors. This method produces realistic and diverse 3D driving scenes from a coarse semantic layout, which to our knowledge, is one of the first of its kind. Finally, we simulate dynamic and realistic driving scenario, compose the 3D world at each time step, and perform stochastic raycasting to produce the final 4D LiDAR sequence. Our generated results are diverse, realistic, temporally coherent, and align with the layout (Fig.~\ref{fig:teaser}).

Our experimental results demonstrate that individual frames generated by LidarDM exhibit \textit{realism} and \textit{diversity}, with performance on-par with state-of-the-art techniques in unconditional single-frame LiDAR point cloud generation.
Moreover, we show that LidarDM can produce \textit{temporally coherent} LiDAR videos, outperforming a robust stable diffusion baseline. We further demonstrate LidarDM's \textit{conditional generation} by showing that the generated LiDAR matches well with ground-truth LiDAR on matching map conditions for both geometry and intensity. Among generative LiDAR methods, LidarDM is the {first to support map-conditioned generation} and the {first to support sequence generation}. Lastly, we illustrate that the data generated by LidarDM exhibit a \textit{minimal domain gap} when tested with perception modules trained on real data and can also be used to augment training data to significantly improve 3D detectors and planners. This gives premise for using generative LiDAR models to create realistic and controllable simulations for training and testing driving models. For detailed applications of LidarDM, please refer to Sec.~\ref{sec:related_works} and Fig.~\ref{fig:apps}.

\section{Related Works}
\label{sec:related_works}

\subsubsection{LiDAR Simulation} 
Realistic LiDAR sensor simulation is crucial for robotics and self-driving vehicle training and testing.  Simulators like CARLA~\cite{carla} and AirSim~\cite{airsim} create environments with static (buildings, trees, street lights) and dynamic (cars, bicycles, buses) assets  and simulate LiDAR with raycasting. Such approaches are simple and easy to integrate, hence are widely used in robot simulation~\cite{macenski2022robot, afzal2021gzscenic}. 
However, these  methods  face limitations in realism and scalability 
due to two key issues: (i) the need for %
3D assets, which are costly and limit variations; (ii)  the sim2real gap for both asset design and physics simulation.

Recent data-driven approaches reconstruct objects and environments from real-world LiDAR as meshes \cite{lidarsim, schmidt2023lidar} or neural fields \cite{nfl,lidarnerf} from which LiDAR are simulated from raycasting or volume rendering. However, these methods require rendering LiDAR from models constructed from real-world scenes, which is costly and not scalable. In contrast, LidarDM creates scenes in a purely generative manner, eliminating the need for man-made or reconstructed assets and environments, and can generate point clouds from environments unseen during training. For example, Champs-Élysées (in Fig.~\ref{fig:apps} (a)) was created from a hand-crafted map layout, which no re-simulation method can achieve.

\subsubsection{LiDAR Generation}
\begin{figure*}[t] \centering  %
 \vspace{0.5em}
\includegraphics[width=0.78\textwidth]{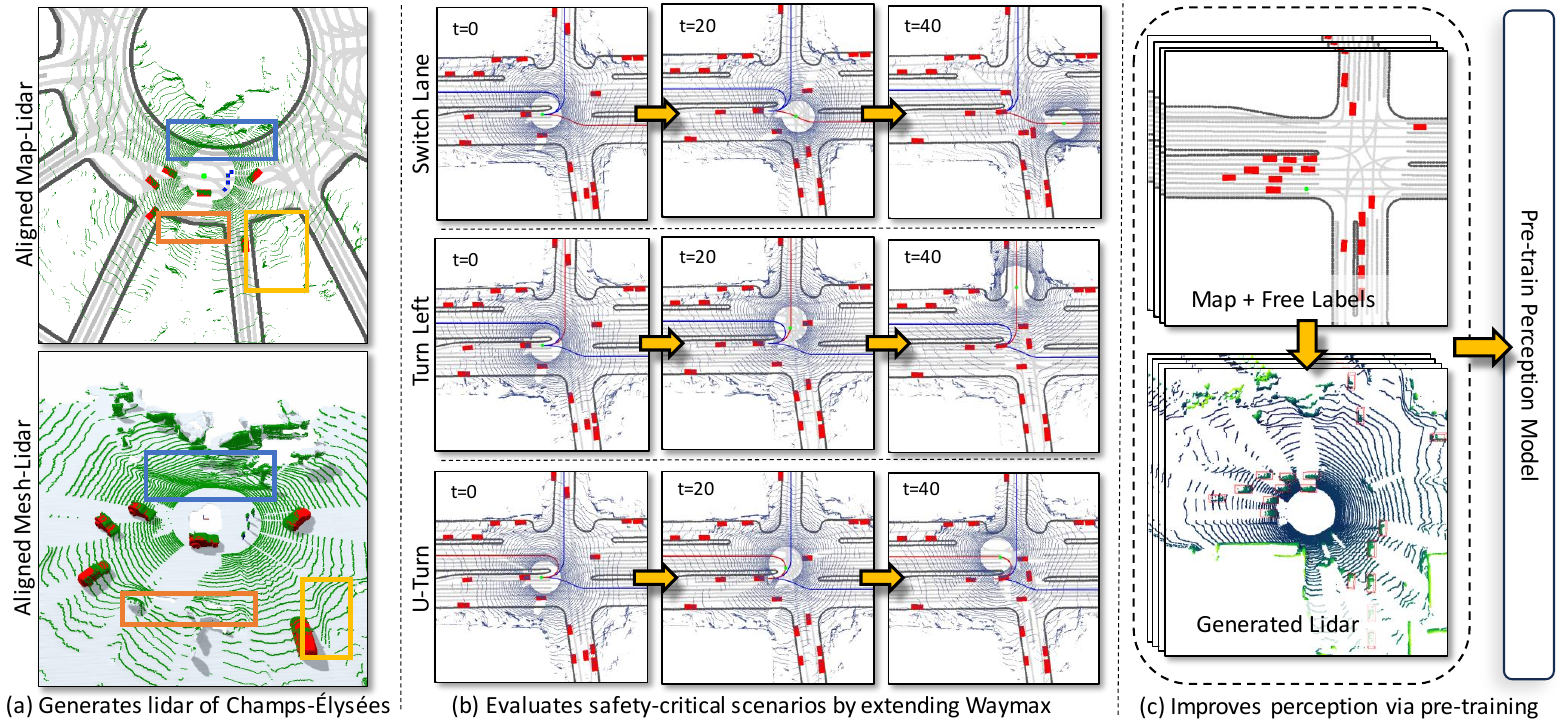}
 \vspace{-0.5em}
 \caption{Applications of LidarDM: (a) generating layout-conditioned LiDAR (color boxes highlight the lidar-map consistency); (b) creating sensor data for traffic simulator \cite{waymax} to enable end-to-end safety-critical scenarios evaluation; (c) generate large volume Lidar data with known ground truth labels to improve perception models without data capturing and labelling. }
 \label{fig:apps}
 \vspace{-2em}
 \end{figure*}

Generative models provide a promising alternative for creating realistic LiDAR point clouds without reconstructing real-world environments. 
Early LiDAR generation works utilized the range image representation using GANs \cite{caccia}, VAEs \cite{caccia}, and diffusion models \cite{lidargen, lidm, rangeldm, r2dm}. Later works used voxel representations with VQGANs \cite{ultralidar}. However, these methods only generate single frame LiDAR, and do not provide controllable or video generation. LidarDM addresses these issues by conditionally generating a 4D world and performing physics-based raycasting. As shown in Fig.\ref{fig:apps} (b) and (c), LidarDM provides realistic LiDAR data to traffic simulators thanks to its temporal consistency and can train perception models (Sec.\ref{sim2real}) with paired semantic layout and generated LiDAR—benefits unmatched by other generative methods.

\section{Layout-Guided LiDAR Video Generation}
\label{sec:method}

Our goal is to create a realistic, physically plausible, and temporally consistent LiDAR sequence that enables a free viewpoint based on a given bird's eye view semantic layout in a purely generative manner without relying on any pre-collected assets like 3D maps.  The key to achieving this lies in first generating and composing the underlying 3D world, followed by using generative simulation to create realistic sensory observations. We begin by formulating generation as a joint 4D scene generation task (Sec.~\ref{sec:formulation}). Next, we discuss leveraging 3D diffusion models to create static and dynamic elements, and ensuring faithful interactions (Sec.~\ref{sec:asset}). Finally, a sensor generation procedure is executed to produce the final LiDAR video (Sec.~\ref{sec:simulation}). Fig.~\ref{fig:overview} depicts the overview of our method.

\subsection{Problem Formulation}
\label{sec:formulation}

Formally, given an input layout \(\mathcal{I} \in \bbR^{L\times W \times M}\) representing traffic elements from a bird's eye view where $L$, $W$, and $M$ are length, width, and map classes (lanes, roads, crosswalks), respectively, our goal is to generate a LiDAR point cloud video \(\mathcal{X} = \{\mathbf{x}_t\}\), with each \(\mathbf{x}_t \in \bbR^{N\times 4}\) being a point cloud (3D location and intensity) at frame \(t\) with \(\mathbf{x}_0\) matching the input layout. This conditional generation setting offers full controllability, and hence lays the foundation for a practical asset-free simulator. Without a map, our approach defaults to unconditional generation (i.e., in Sec. \ref{sec:exp_uncond_gen}).

\subsubsection{4D World Representation} Our key technical innovation to address the challenge lies in jointly modeling the generation of underlying 4D world together with sensor generation. We define the world scene representation as \(\mathcal{W} = \{ \mathbf{s}, \{\mathbf{o}_i\}_{i=0}^N\}\), where \(\mathbf{s}\) represents a static scene geometry and \(\mathbf{o}_0, ..., \mathbf{o}_N\) are dynamic objects. Both are represented in the form of an occupancy grid. 
To model dynamics, we additionally consider the actions of these dynamic objects in the form of trajectories \(\mathcal{P} = \{ \boldsymbol{\tau}_{0},..., \boldsymbol{\tau}_{T} \}\), with \(\boldsymbol{\tau}_t = \{ \boldsymbol{\xi}_\mathrm{ego}, \{\boldsymbol{\xi}_{i, t}\}_{i=0}^N \}\) representing the pose of actor \(i\) at time \(t\) as well as egocar pose $\boldsymbol{\xi}_\mathrm{ego}$. The pose for rigid objects and the egocar lies in the \(\mathbb{SE}(3)\) space, while for articulated objects like pedestrians, it is represented as a kinematic chain. 
A composed scene represent the states of the world at $t$, incorporating the poses of the ego car and dynamic objects at time \(t\), is denoted by \(\mathcal{W}_t = \pi(\mathcal{W}, \boldsymbol{\tau}_t)\), where \(\pi\) is an operator composing actors to world.

\subsubsection{4D World and LiDAR Generation} 
To ensure realism and consistency over time and between the world and sensory readings, we formulate the generation task as a sampling problem from the joint distribution \(p(\mathcal{X}, \mathcal{P}, \mathcal{W} | \mathcal{I})\). Directly modeling and sampling the joint distribution, however, is challenging as it involves estimating a distribution across multiple data modalities (e.g., car trajectories, scene layouts, sensor noise, etc.). To tackle this, we factorize the joint distribution \(p(\mathcal{X}, \mathcal{P}, \mathcal{W} | \mathcal{I})\) as follows:
\[
\underbrace{p(\mathbf{s} | \mathcal{I}) \cdot 
\prod_i p(\mathbf{o}_i | \mathcal{I})}_{\text{3D scene and object gen}} \cdot 
\underbrace{\prod_t p(\boldsymbol{\tau}_t | \boldsymbol{\tau}_{<t}, \mathcal{W}, \mathcal{I})}_{\text{trajectory gen}} \cdot 
\underbrace{\prod_t p(\mathbf{x}_t | \boldsymbol{\tau}_t, \mathcal{W})}_{\text{sensor simulation}}.
\]
Next, we will discuss each individual task in detail.

\begin{figure*}[t] 
\centering  %
\vspace{0.5em}
\includegraphics[width=0.9\textwidth]{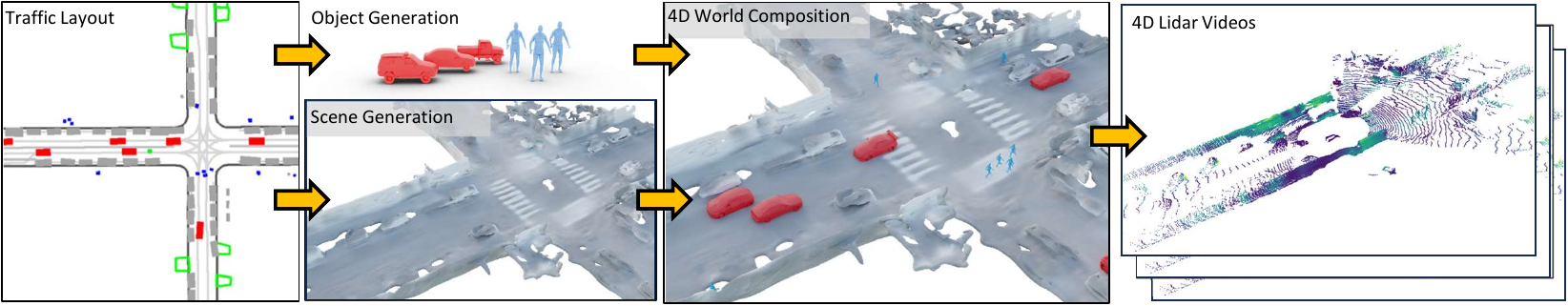}
 \caption{Overview of the LidarDM: Given the input traffic layout at time $t=0$, \method~begins by generating actors and the intensity-infused static scene. We then generate the motion of the actors and the egocar, and compose the underlying 4D world. Finally, a generative- and physics-based simulation is used to create realistic 4D sensor data.}
 \label{fig:overview}
 \vspace{-2em}
 \end{figure*}

\subsection{Scene, Object and Trajectory Generation} 
\label{sec:asset}

We decompose the world into a static, intensity-infused background scene that is constant over time and dynamic foreground objects that move. This decomposition simplifies the challenging 4D world generation into manageable tasks: creating object geometries and generating dynamic effects, while ensuring temporal consistency (\eg constant cars' shapes and walls and trees over time) and physical plausibility (\eg ensuring correct collision reasoning).

\subsubsection{Scene Generation} 
The scene generation step addresses the problem of sampling the geometry and LiDAR intensity of a scene from a given input layout \(\mathcal{I}\): \(\mathbf{s} \sim p(\mathbf{s} | \mathcal{I} )\). We parameterize the intensity-infused scene  \(\mathbf{s} \in \mathbb{R}^{2 \times L \times W \times H}\), where each entry \(s_j \in \mathbb{R}^2\)  encodes the truncated signed distance to the surface and an intensity value. %

{\it Model:}~We leverage the latent diffusion model~\cite{stablediffusion, stablediffusion} to tackle this challenge of modeling and sampling from $p(\mathbf{s} | \mathcal{I})$ due to its capacity to sample high-quality data while effectively incorporating strong conditional guidance. Specifically, our model encodes the high-dimensional $\bs$ into a continuous latent representation $\bz$ using an encoder-decoder structure~\cite{stablediffusion} with a scene encoder $E_\theta(\bs) = \bz$ and a scene decoder $D_\theta(\bz) = \Tilde{\bs}$. 
This encoder-decoder structure efficiently compresses the input data into a lower-dimensional latent space, enabling more effective and efficient sampling. Additionally, we encode our high-definition map layout \(\mathcal{I}\) into a latent space \(\mathbf{c} = M_\theta (\mathcal{I})\), allowing for more compact conditioning. 

{\it Sampling:}~We leverage a probabilistic denoising diffusion model~\cite{stablediffusion, ncsn} $F_\theta(\bz, \bc)$ to perform classifier-free guidance sampling~\cite{classifierfreeguidance}.
Specifically for each diffusion step $k$, the following Langevin dynamics step is performed to progressively denoise until a clean sample $\bz_0$ is acquired: 
\[
    \mathbf{z}_{k-1} = \mathbf{z}_{k} + \frac{\lambda_k}{2} \left[ (1+w)F_\theta(\mathbf{z}_k, \mathbf{c}) - w F_\theta(\mathbf{z}_k) \right] + \sqrt{\lambda_k} \boldsymbol{\epsilon}_k
\]
$F_\theta(\mathbf{z}_k, \mathbf{c})$ is the score function $\nabla_\bz \log p(\bz | \bc)$ of the conditional distribution at $\bz_k$ and $F_\theta(\mathbf{z}_k) = F_\theta(\mathbf{z}_k, \bc=\mathbf{0})$ is the the score function of the unconditional distribution $p_\theta(\bz)$. $w$ is the CFG guidance scale parameter, $\lambda_k$ is an annealed noise schedule parameter, and $\boldsymbol{\epsilon}_k \sim \mathcal{N} (\mathbf{0}, \bI)$. 
Finally, a 3D scene sample $\bs$ is recovered by decoding the reverse-diffused sampling latent code $\bs = E_\theta (\mathbf{z}_0)$. %

{\it Training:} We train our diffusion-based scene generation model using a dataset that pairs scene geometry with map conditioning. Direct access to dense scene geometry is not available in practice. Instead, we use NKSR~\cite{nksr}, a textured 3D reconstruction method, to recover a pseudo-GT from an input LiDAR sequence. The recovered mesh's texture encodes intensity. Ground truth annotations are used to remove moving dynamic objects, ensuring our reconstruction contains only the static scene and objects. We then train the auto-encoders for both scene geometry and map layout using reconstruction loss and KL divergence loss: \(\min_\theta \mathcal{L}_\mathrm{recon} + \mathcal{L}_\mathrm{KL}\) over real-world examples. Our latent diffusion model is trained using the score matching loss function: \(\mathcal{L}_\mathrm{LDM} = \mathbb{E}_{(\bz, \bc), \boldsymbol{\epsilon}, k} \left[ \| \boldsymbol{\epsilon} - F_\theta(\bz_k, k, \bc)\|_2^2\right]\), where \(\bz_k\) is the forward diffused noisy sample at step \(k\) .

\subsubsection{Object Generation}
We employ two object generation frameworks, GET3D \cite{get3d} and AvatarClip \cite{avatarclip}, to create dynamic traffic participants. For each actor \(\mathbf{o}_i\) in a given layout \(\mathcal{I}\), we sample a random variable \(\mathbf{z} \sim \mathcal{N}\) and generate the corresponding actor mesh following \(\mathbf{o}_i = G(\mathbf{z})\), where \(G(\cdot)\) is the generator/decoder of the chosen generative method.  We use GET3D to generate vehicles and then rescale them to properly fit within a target bounding box of the layout. For pedestrian generation, we utilize AvatarClip \cite{avatarclip}, which is conditioned on a SMPL \cite{SMPL} pose and shape parameter \(\bf{p}=(\boldsymbol{\theta}, \boldsymbol{\beta})\). We use Mixamo \cite{mixamo} to animate the rigged model, ensuring realistic 4D human walking motion. %

Together, the generated static world \(\mathbf{s}\) and each actor \(\mathbf{o}_i\) define our 3D world, denoted as \(\mathcal{W}\), as depicted in Fig.~\ref{fig:overview}

\subsubsection{Trajectory Generation}

We extend Waymax \cite{waymax}, a data-driven 2D BEV traffic simulator, to control the behaviors of traffic actors in more systematic manners. Given a scenario from the WOMD Dataset \cite{womd}, we use Waymax to replay ego-vehicle's and agent's real-world trajectories, with an additional reactive Intelligent Driver Model \cite{idm} that updates each agent's acceleration to avoid collisions. For unconditional generation, we sample trajectories from a trajectory bank obtained from Waymo Open dataset \cite{waymoopen}. We employ heuristics to ensure physical feasibility (no hovering) and that no collisions occur (through agent-agent or agent-scene collisions). Around \textbf{12.3\%} of sampled trajectories are retained, which is acceptable because resampling is trivial.

This approach renders our world generation to be completely asset-free, end-to-end generative, and temporally consistent, allowing for a realistic and physics-based simulation without the need for artist-curated~\cite{carla} or pre-collected assets~\cite{lidarsim, unisim} as in previous LiDAR simulation methods.

\subsection{Physics-Informed LiDAR Generation}
\label{sec:simulation}

Given the complete 4D world \(\mathcal{W}\) and the poses \(\mathcal{P}\), our next step is to generate a realistic LiDAR point cloud corresponding to these conditions. At a high level, we use the poses to compose the scene and objects at each timestep, then perform physics-informed ray casting to obtain purely physically simulated LiDAR as an intermediate result. We leverage data-driven conditional sampling to {generate the final point cloud to} simulate real-world LiDAR noises. %

\subsubsection{Scene Composition}  
We use Dual Marching Cube ~\cite{dmc} to obtain the 3D mesh of the static world from the  TSDF channel of the generated \(\bs\). Then, using our generated actor trajectories, we transform all 3D agents' meshes to the world coordinates and compose it with \(\bs\) using ego poses and all actors actor poses at time \(t\), \(\tau_t\), producing the full world geometry at each time \(t\):
$
\mathcal{W}_t = \pi(\mathcal{W}, \boldsymbol{\tau}_t)
$.
For vehicles, \(\pi\) applies a rigid transform. For pedestrians, $\pi$ applies a rigid transform and articulates the human body shape to simulate animated movement with forward kinematics~\cite{bogo2016keep}. 

\subsubsection{Physics-based Ray Casting}

LiDAR sensors acquire a 3D point cloud by shooting beams of light from the sensor into the scene. Raycasting simulates this process for a single beam by calculating the $\{x,y,z\}$ coordinate point where a beam would intersect with the scene based on a given ego vehicle position $\tau_t$, the composed scene $\mathcal{W}_t$, and the beam's elevation ($\theta$) and azimuth ($\phi$) angles. LiDARs typically have multiple beams that spin to generate a single 3D point cloud. Therefore, the process is repeated for each beam based on the virtual sensor configuration 
which we match with the real-world LiDARs based on provided KITTI HDL-64E \cite{vlp} and Waymo specifications.

\subsubsection{Intensity Modeling}
\label{sec:intensity}
For each point $\overline{\mathbf{x}}^s_{it} \in \overline{\mathbf{x}}_t$ that lies on the background scene, we query the intensity channel of the volume \(\bs\) for the intensity of the closest vertex. 
For intensity values of dynamic objects, we opt for a physical-based approach following Lambert's Cosine Law. Namely, for each point $\overline{\mathbf{x}}^o_{it} \in \overline{\mathbf{x}}_t$ that lies on the dynamic object, $intensity(\overline{\mathbf{x}}^o_{it}) = \alpha + \beta \left(I_{it} \cdot N^o_{t}\right) $ where $I_{it}$ denotes the vector from the lidar sensor to $\overline{\mathbf{x}}^o_{it}$, $N^o_{t}$ denotes the normal vector of the object at $\overline{\mathbf{x}}^o_{it}$, and $\alpha, \beta$ are constants that have been hand-tuned to match the empirical intensity distribution of dynamic objects in Waymo Dataset.

\subsubsection{Stochastic Raydrop}

Raycasted LiDAR from the generated world appears overly clean, without real-world environmental and sensor noise. Inspired by LiDARSim, we stochastically simulates ``raydrop'', where rays do not return to the sensor.  For each raycast scan at time \(t\), \(\overline{\mathbf{x}}_t\), we project it onto a 2D spherical range image and predict raydrop probability per pixel on this image using a U-Net architecture~\cite{rangenet}, supervised by real-world LiDAR scan raydrop masks. Our approach, unlike LiDARSim, requires only a range map, eliminating the need for multiple additional metadata input channels that are only available in real-world data. We sample from this mask with a Gumbel sigmoid to produce the final LiDAR scan \(\mathbf{x}_t\) for each frame, concluding the end-to-end LiDAR video generation process.

\section{Experiment}

 \begin{figure*}[t] \centering  %
 \addtolength{\tabcolsep}{-0.2em}
 \begin{tabular}{cccccc} \centering
 \includegraphics[width=0.12\textwidth]{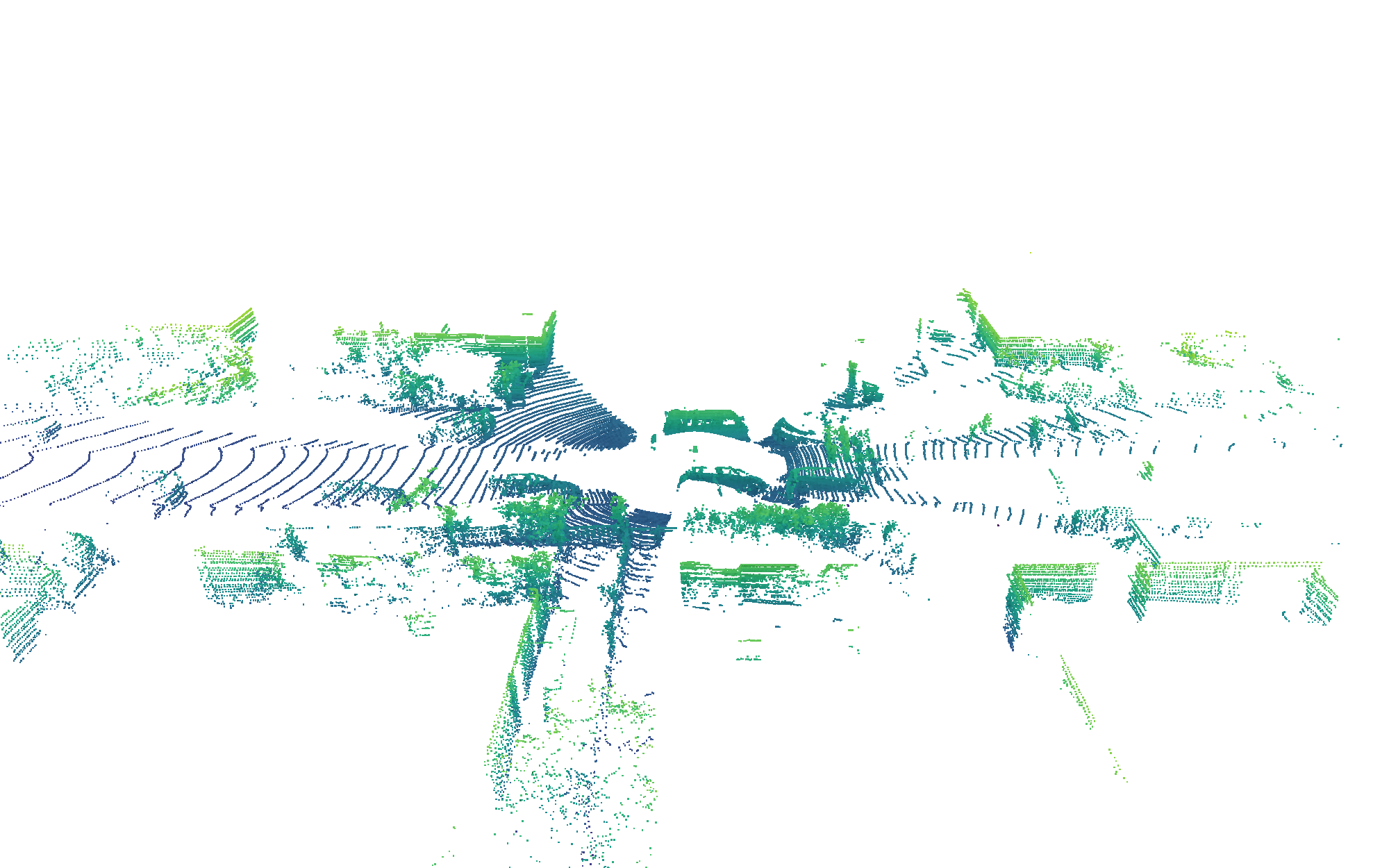}
 &\includegraphics[width=0.12\textwidth]{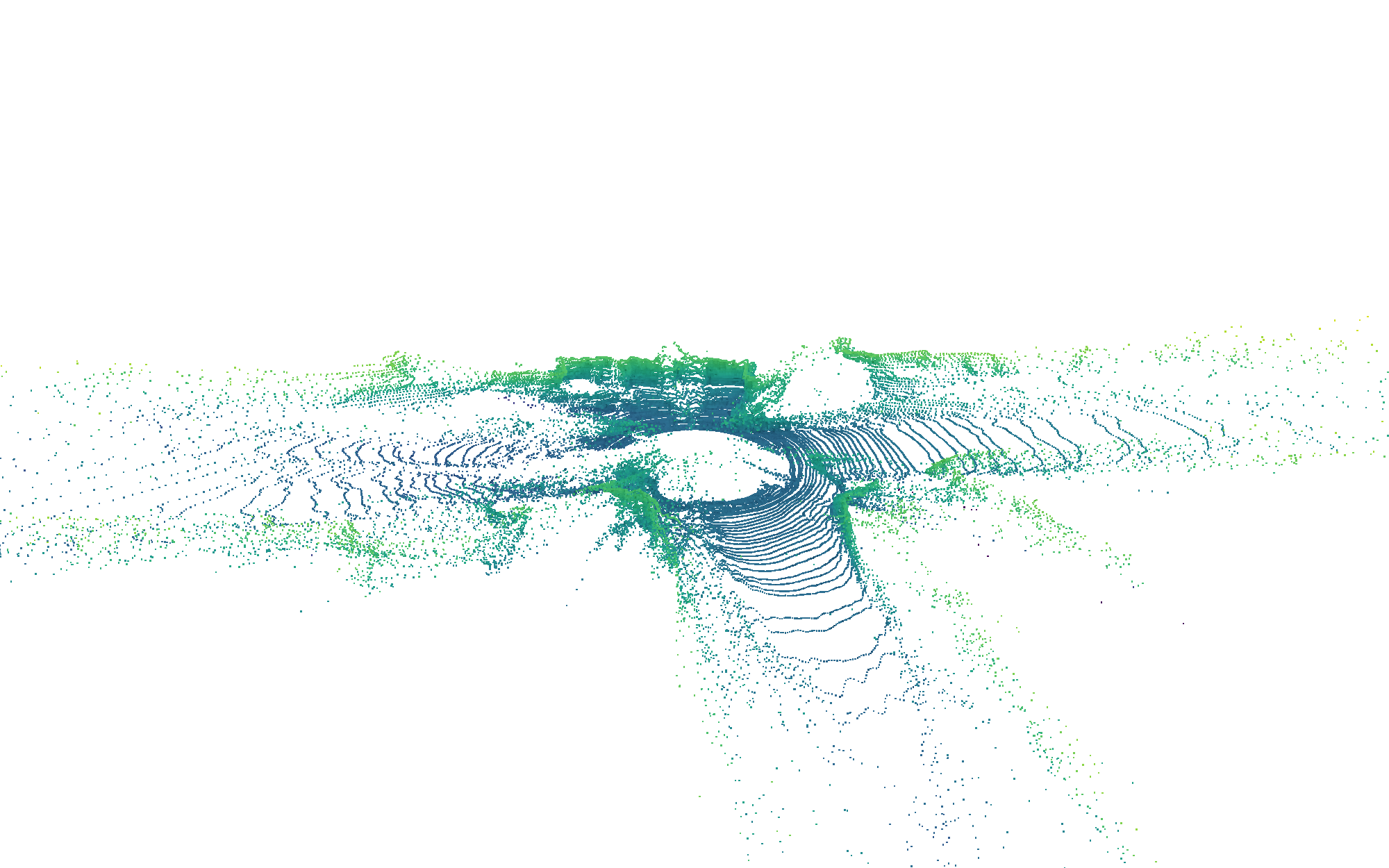}
 &\includegraphics[width=0.12\textwidth]{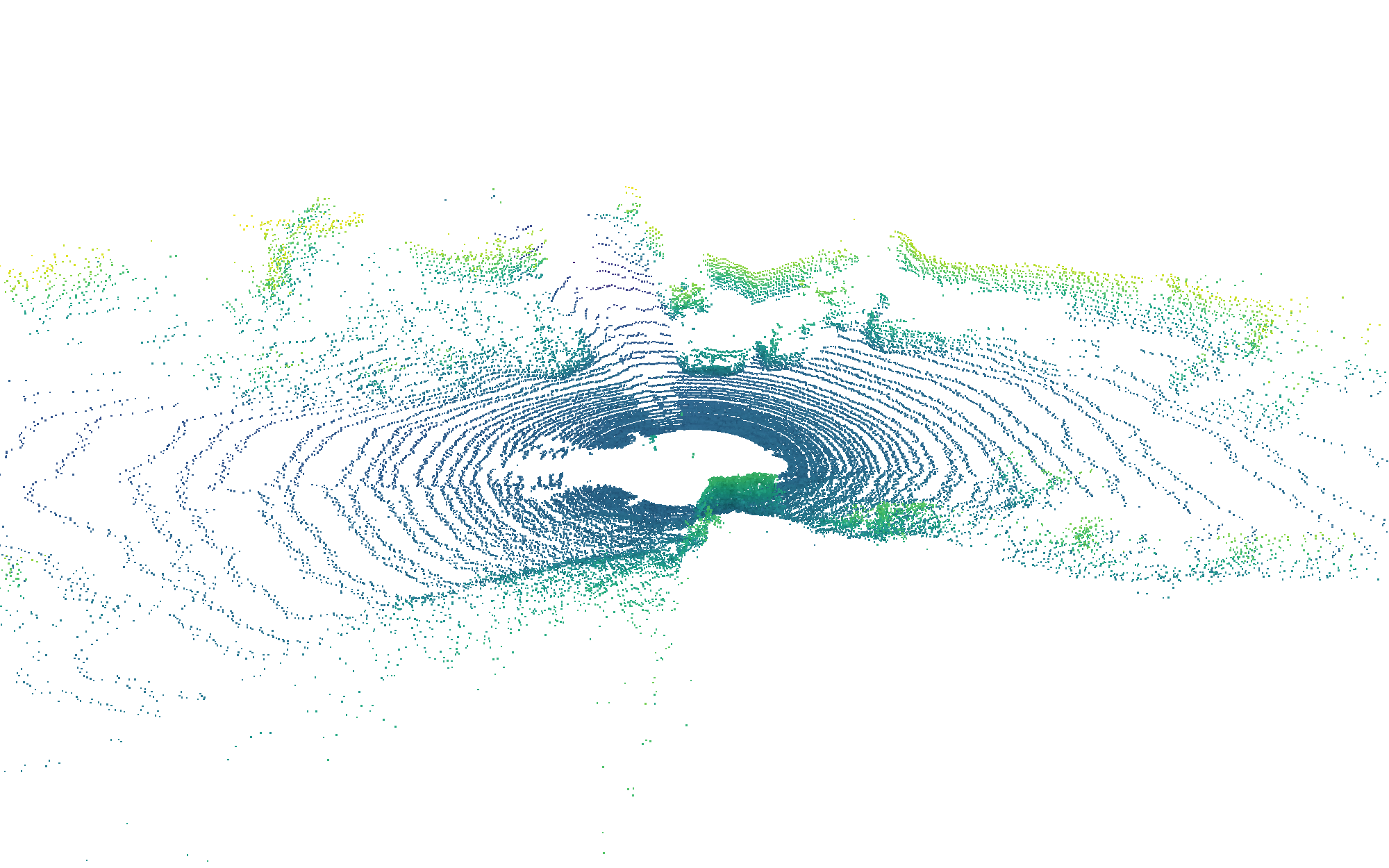}
&\includegraphics[width=0.12\textwidth]{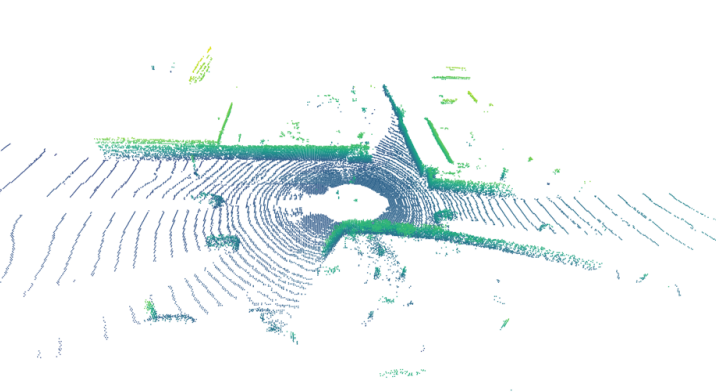}
&\includegraphics[width=0.12\textwidth]{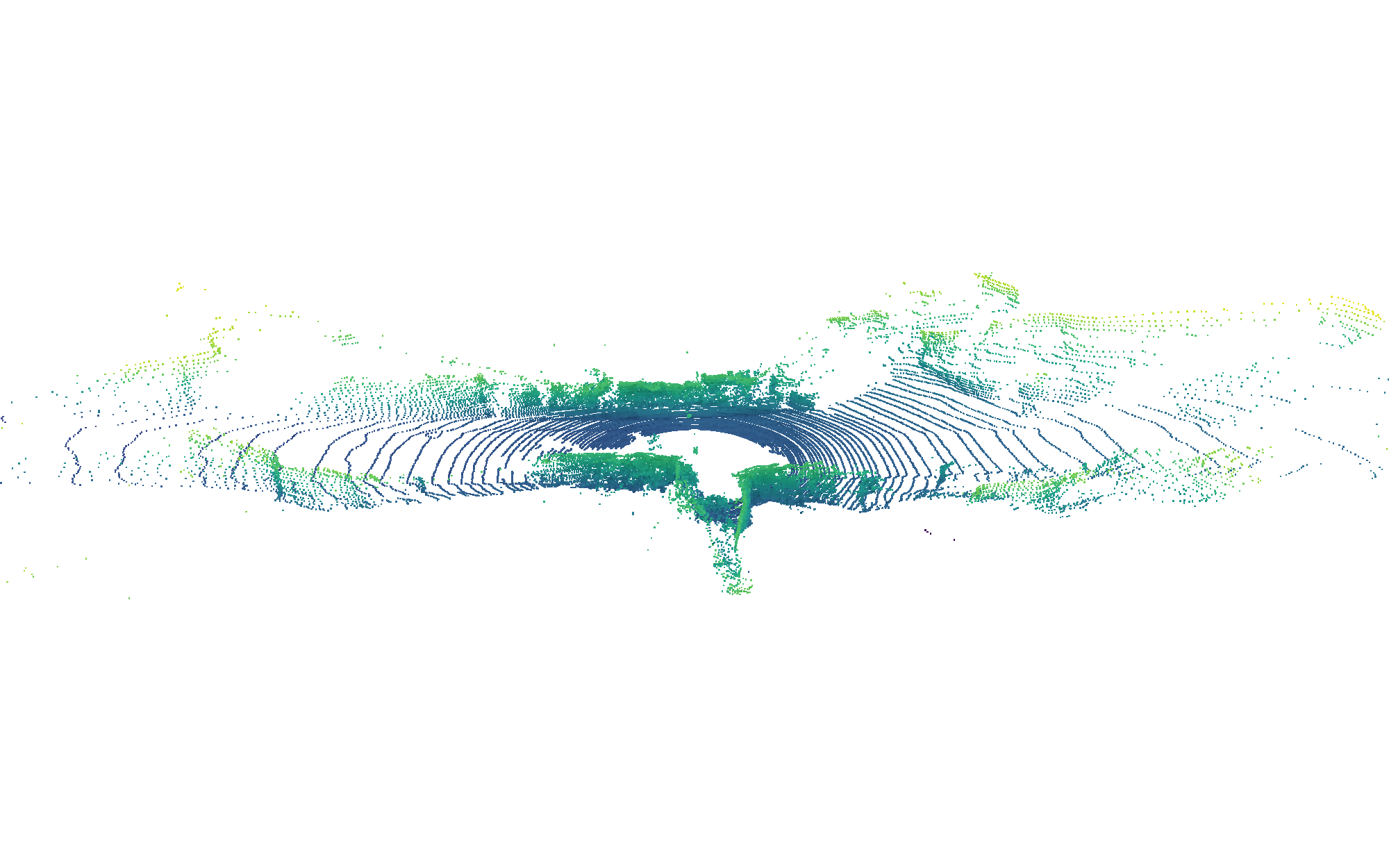}
&\includegraphics[width=0.12\textwidth]{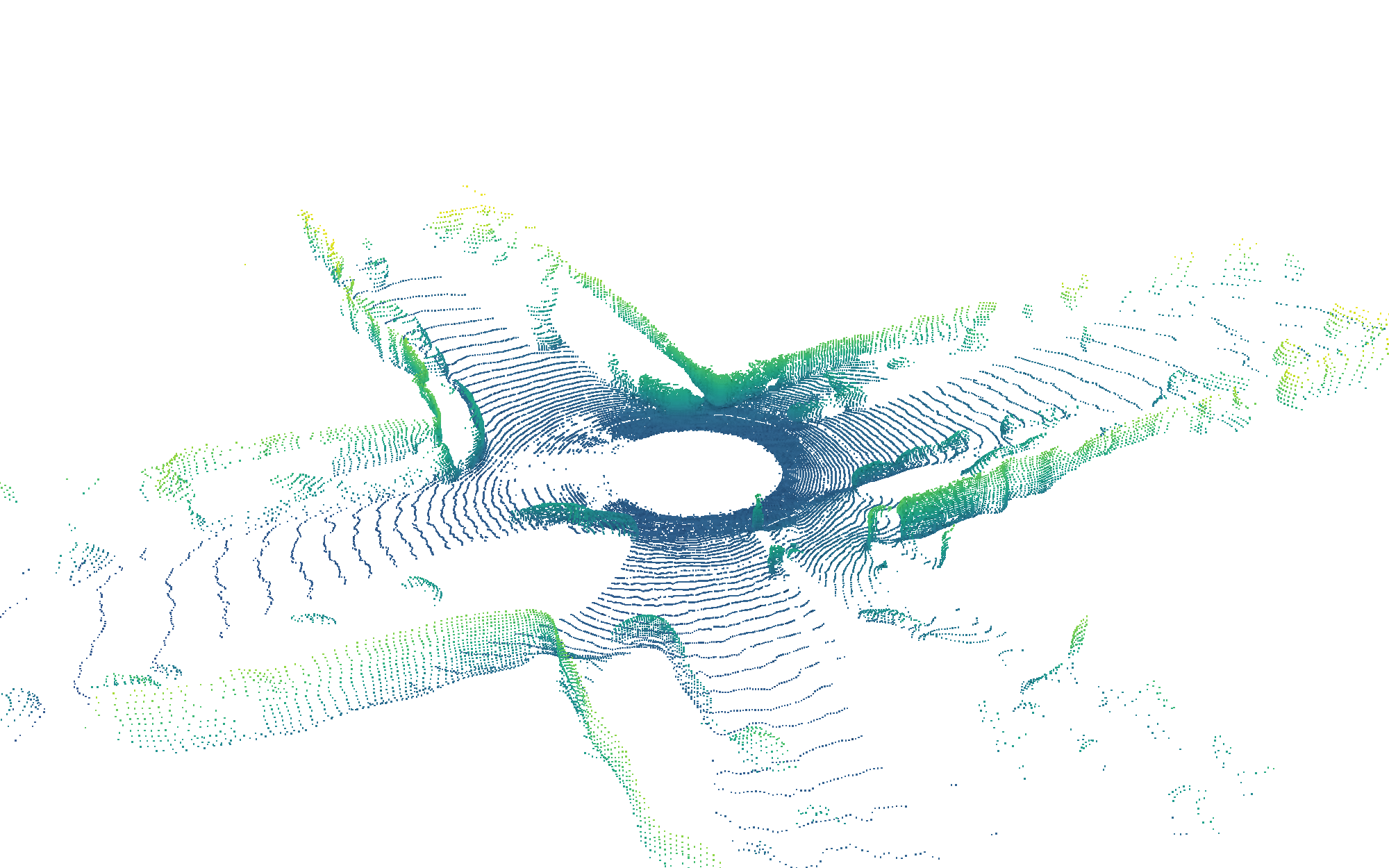} 
\\

\includegraphics[width=0.12\textwidth]{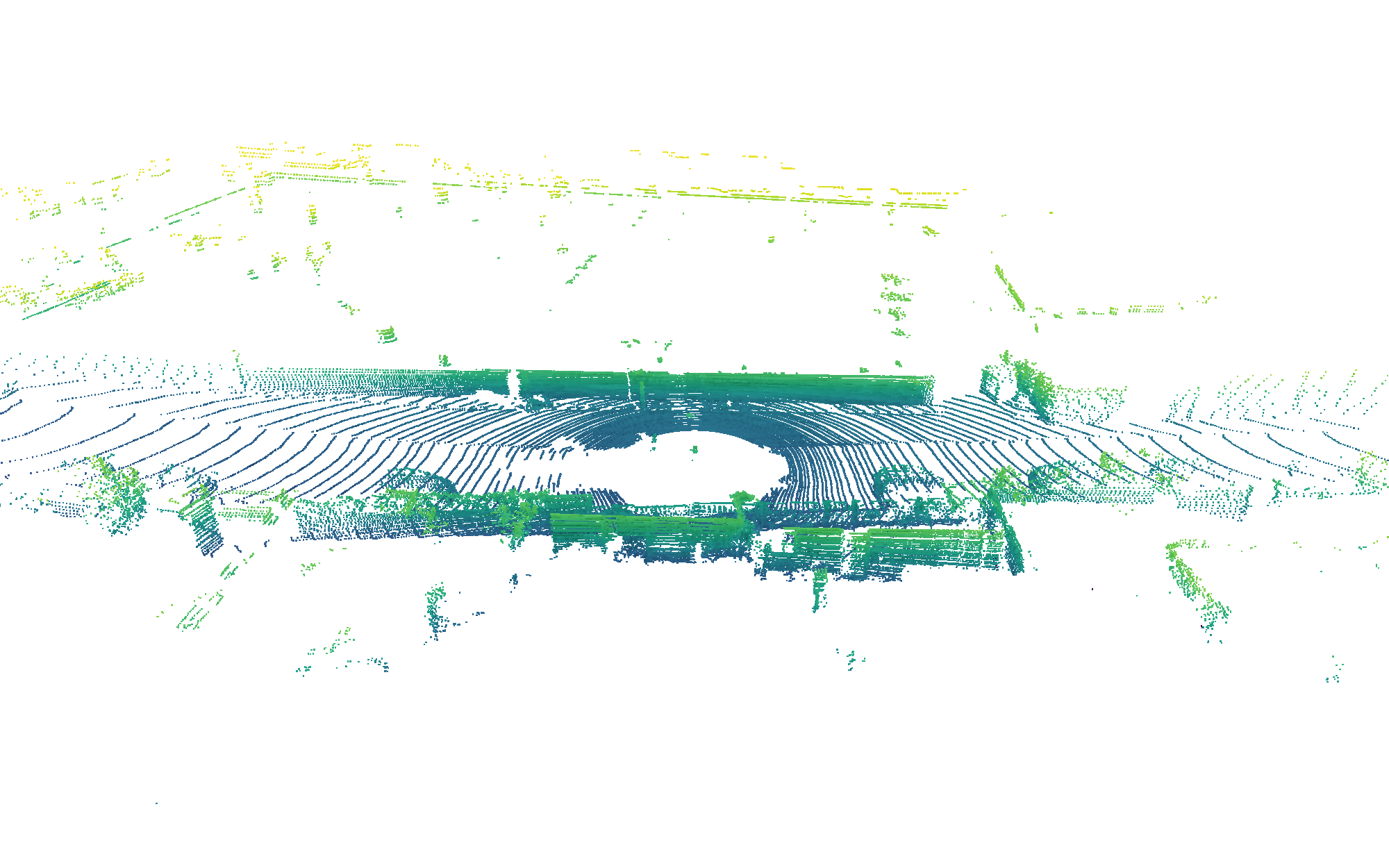}
 &\includegraphics[width=0.12\textwidth]{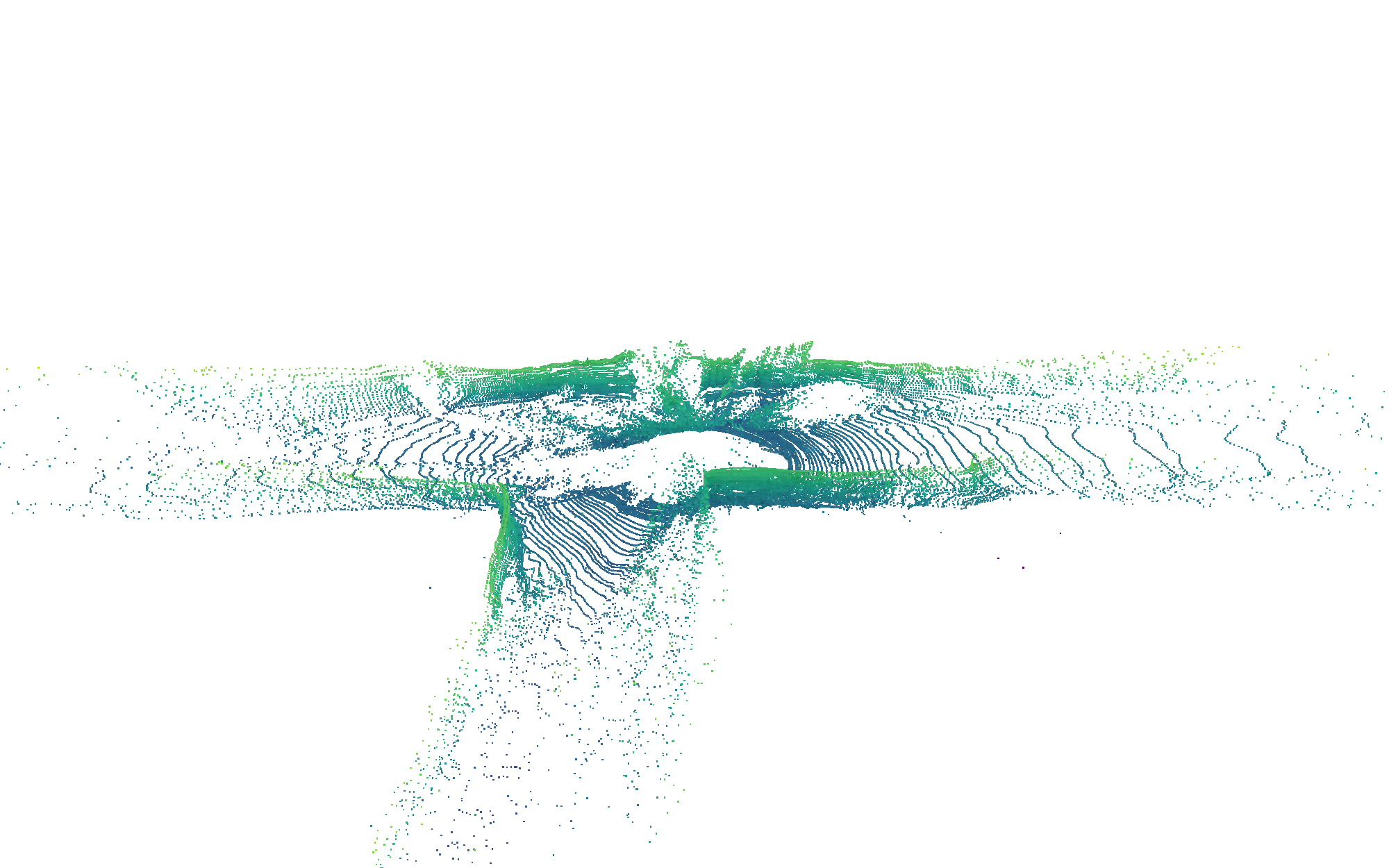}
 &\includegraphics[width=0.12\textwidth]{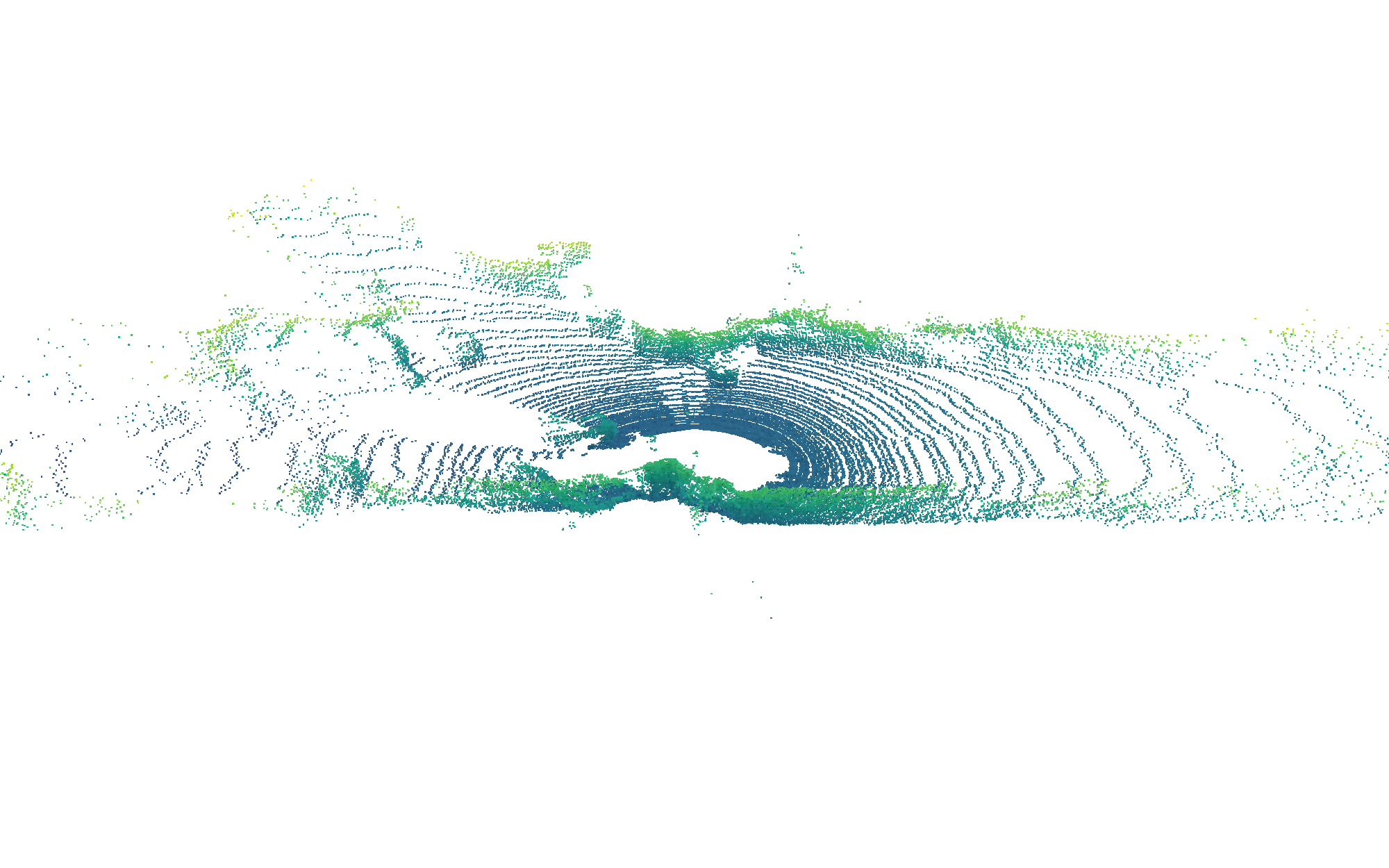}
&\includegraphics[width=0.12\textwidth]{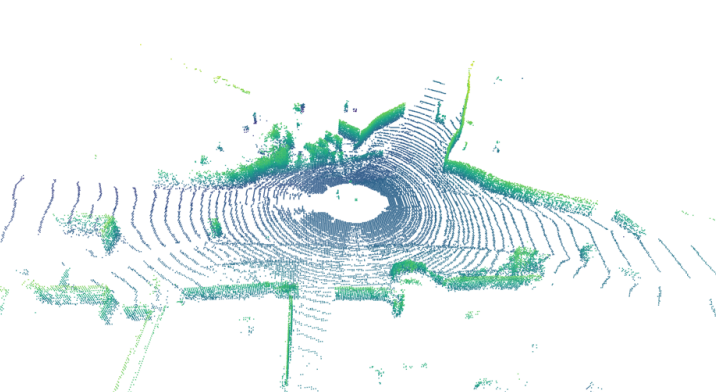}
&\includegraphics[width=0.12\textwidth]{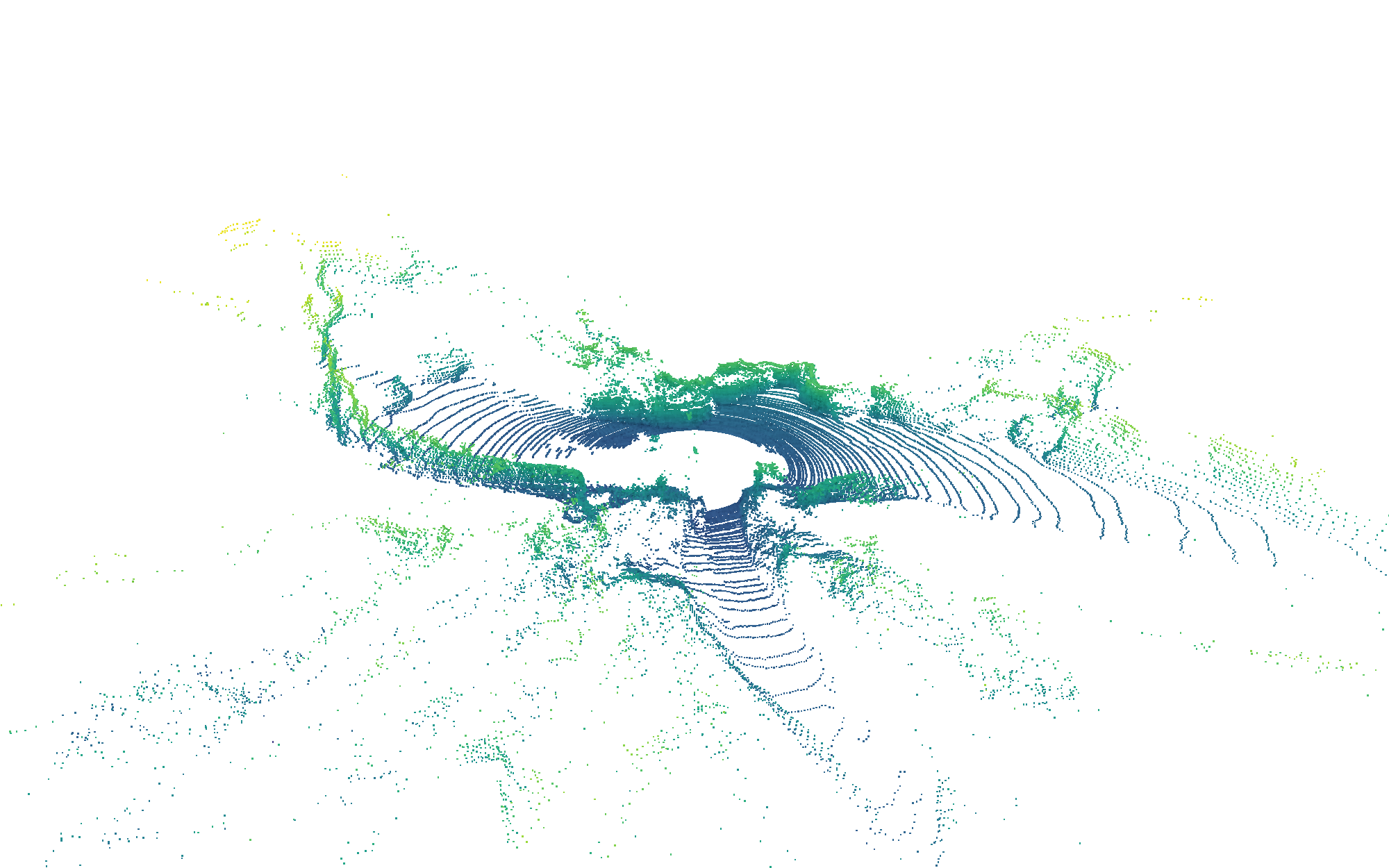}
&\includegraphics[width=0.12\textwidth]{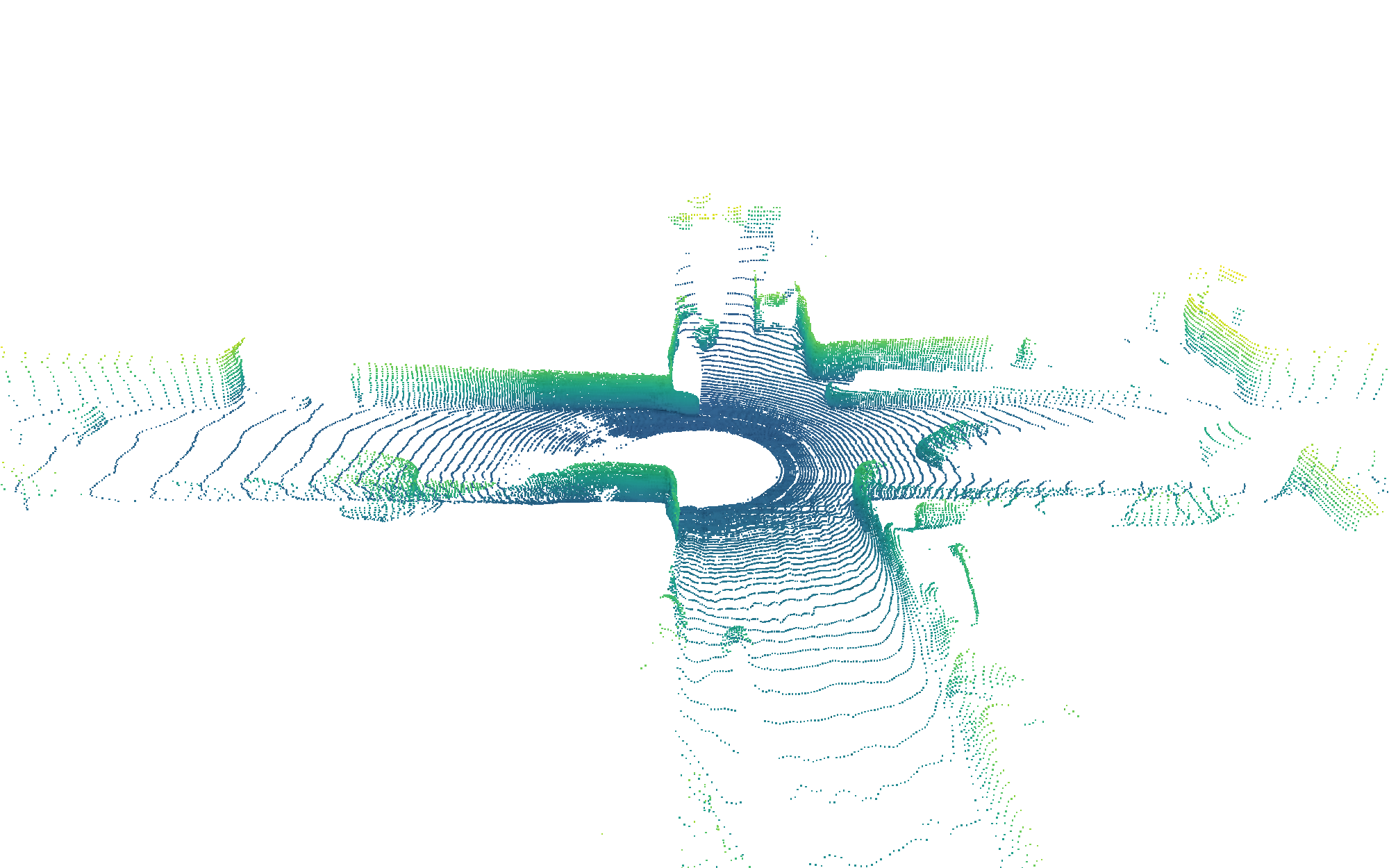}
\\

Real & ProjectedGAN & LiDARGen & UltraLiDAR & R2DM & Ours 
\end{tabular}
 \caption{Real KITTI-360 samples vs unconditional samples from the competing methods. LidarDM generates samples that feature more detailed salient objects (e.g., cars, pedestrians), sharper 3D structures (e.g., walls), and realistic road layouts.}
 \label{fig:kitti_uncond_comparison}
 \vspace{-2em}
 \end{figure*}

\subsection{Setup}

\subsubsection{Datasets} We evaluate LidarDM on the KITTI-360 \cite{kitti360} and Waymo Open \cite{waymoopen} datasets. 
KITTI-360 contains nine driving sequences (76,715 samples), where the first sequence is used as a val sequence (11,518 samples) and the last eight are used for training (65,197 samples).  However, KITTI-360 does not provide detailed BEV HD map information limiting its applications in conditional models. Waymo Open \cite{waymoopen} is a dataset containing 1048 sequences with 158,000 training and 29,700 validation frames. 
The dataset provides an HD map in a vector format which we rasterize into a segmentation map and use for training the conditional model.
The dimensions of the map tensor are \texttt{L}$\times$\texttt{W}$\times$\texttt{M} (length$\times$width$\times$map classes). The map has 5 classes (lane markings, road lines, edges, crosswalks, driveways).

\subsubsection{Training Details} We train our models for 48 hours using four Nvidia A100 40GB GPUs. We use the Adam optimizer with a learning rate of 1e-4 for the VAE and 1e-5 for the diffusion U-Net, with a cosine decay schedule. 

\subsubsection{Model Details} 
The latent diffusion is a UNet with 5 ResNet blocks, featuring channels of {128, 128, 256, 512, 512}.  The SDF VAE has 4 ResNet blocks with channels of {448, 640, 896, 1280}. 
The Map VAE has ResNet down/upsample blocks with channels of {64, 64, 128, 256, 512}.

\subsection{Baselines}

\subsubsection{Unconditional Generation} LiDARVAE, LiDARGAN, ProjectedGAN, LiDARGen, and R2DM are baselines that use range image representation whereas UltraLiDAR uses BEV voxel space. For fair comparison, we follow UltraLiDAR and evaluate MMD and JSD on a histogram of voxel occupancy instead of voxel density \cite{ultralidar}.\footnote{
LiDM has random yaw rotations in their samples that can cause unfair comparison (confirmed with authors). We will include once they provide rotation-free samples}%

\subsubsection{Temporal Coherency} We are the first to attempt the task of sequential LiDAR generation and thus no previous models exist for comparison. Nonetheless, we implement a sequence diffusion baseline inspired by recent work in video generation. Concretely, we train a VAE to encode individual LiDAR frames. This has been shown previously \cite{ultralidar} to be effective. Next, we train a diffusion model to directly denoise multiple (\ie, 5) LiDAR frames at once. %

\subsection{Unconditional Single-Frame Generation}
\label{sec:exp_uncond_gen}

We first validate our model architecture design and showcase our model's generative capability by directly comparing against previous LiDAR generation models in unconditional generation (without using HD maps) on KITTI-360 \cite{kitti360}. 
Based on the results in Table \ref{tab:kitti_uncond}, BEV models (ours or UltraLiDAR) perform best  compared to range image models. Note that UltraLiDAR was directly trained on the task of modeling single LiDAR scans which the benchmark evaluates, which explains the performance gap compared to ours. We also show qualitative comparisons against the baselines in Fig. \ref{fig:kitti_uncond_comparison}.

 \begin{table}%
 \setlength{\tabcolsep}{2pt}
 \small\centering
 \vspace{2mm}
 \begin{tabular}{lccccc}
 \toprule    
  Method & Int. & Con. & Tem.  & MMD$_{\texttt{BEV}}$ ($\downarrow$) & JSD$_{\texttt{BEV}}$ ($\downarrow$) \\
 \midrule
 LiDAR VAE \cite{caccia} & \xmark & \xmark & \xmark & $8.53$e$-4$ & $0.267$ \\
 LiDAR GAN \cite{caccia} & \xmark & \xmark & \xmark & $8.95$e$-4$ & $0.243$ \\ 
 ProjectedGAN \cite{projectedgan} & \xmark & \xmark & \xmark & $7.07$e$-4$ & $0.201$ \\ 
 LidarGen \cite{lidargen} & \cmark & \xmark & \xmark & \cellcolor{bronze}$2.95$e$-4$ & \cellcolor{bronze}$0.136$ \\ 
 UltraLidar \cite{ultralidar}  & \xmark & \xmark & \xmark
 & \cellcolor{gold}{$9.67$e$-5$} & \cellcolor{silver}$0.132$ \\ 
 R2DM \cite{r2dm}  & \cmark & \xmark & \xmark
 & $3.60$e$-4$ & $0.148$ \\ 

  \midrule
 LidarDM (Ours) & \cmark & \cmark & \cmark & \cellcolor{silver} {$1.67$e$-4$} & \cellcolor{gold} {$\mathbf{0.119}$}\\ 
 \bottomrule
 \end{tabular}
 \caption{{Qualitative results for unconditional generation on KITTI-360 dataset. (Int: Intensity, Con: Controllability, Tem: Temporal Consistency) ($\fboxsep=0pt\fbox{\color{gold}\rule{2.5mm}{2.5mm}}$ best, 
 $\fboxsep=0pt\fbox{\color{silver}\rule{2.5mm}{2.5mm}}$ second best, 
 $\fboxsep=0pt\fbox{\color{bronze}\rule{2.5mm}{2.5mm}}$ third best) }}
 \label{tab:kitti_uncond}
 \vspace{-3mm}
 \end{table}

\subsection{Map-conditioned Multi-Frame Generation}
\label{sec:exp_cond_gen}

Our model is the first fully generative LiDAR model that can generate controllable (through map conditioning), realistic, and temporally coherent synthetic LiDAR scans. We will then validate these properties in this section.

\subsubsection{Consistent 4D Generation}

\begin{table} %
\centering
\setlength{\tabcolsep}{4pt}
\small\centering
\begin{tabular}{lcc}
\toprule    
 Metrics & Sequence Diffusion & LidarDM \\
\midrule
\makecell[l]{Total ICP Energy [m] ($\downarrow$)} & 3616.58 & 916.94 \\  
\makecell[l]{Average ICP Energy ($\downarrow$)} & 0.078 & 0.014 \\  
\makecell[l]{Outlier Percentage ($\downarrow$)} & 20.56\% & 7.12\% \\  
\makecell[l]{Chamfer Distance [m] ($\downarrow$)} & 0.39 & 0.17 \\  
\bottomrule
\end{tabular}
\caption{Temporal consistency. Outlier percentage uses distance threshold $\tau = 0.5m$. }
\label{tab:consistency}
\vspace{-7mm}
\end{table}

One of our key contributions is the temporal consistency of the sequential LiDAR generation. To evaluate this, we first use ICP alignment to calculate a relative transformation between consecutive generated frames. We define an \textbf{average point-to-plane energy} over a sequence of LiDAR scans as our quantitative metrics, following this equation:
$
    E = \dfrac{1}{T}\sum_{t=1}^T \texttt{point2plane}(x_{t}, x_{t-1})
$
where $\texttt{point2plane}$ represents the point-to-plane distance~\cite{p2p}, and $x_t$ indicates the LiDAR scan at time $t$. Intuitively, $E$ is prone to higher values from dynamic objects, but it is still a valuable metric to determine if the general scene geometry is preserved over time. To further evaluate the geometric consistency , we also measure the \textbf{outlier point ratio}, defined as the percentage of points with the $\texttt{point2plane}$ distance larger than a certain threshold $\tau$. Table \ref{tab:consistency} shows our quantitative results, where we beat the baseline in both metrics by a notable margin, clearly demonstrating LidarDM's temporal consistency. %

 \begin{figure}[t] \centering  %
 \vspace{0.25em}
 \includegraphics[width=\linewidth]{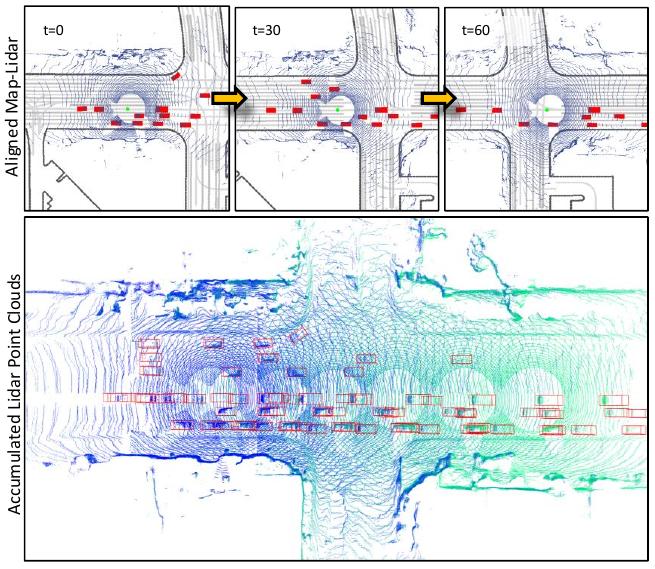}
 \caption{Qualitative results of map-conditioned generation. Accumulated LiDAR shows temporal and layout consistency.
 }
 \label{fig:waymo_cond_qual}
 \vspace{-2em}
 \end{figure}

\subsubsection{Layout-aware LiDAR Generation}

To ascertain the layout-awareness of our LidarDM, we use CenterPoint~\cite{centerpoint} trained on real-world LiDAR scans to validate whether it can accurately detect objects from the LidarDM's LiDAR scan. %

Given an input layout $\mathcal{I}$, we generate the corresponding LiDAR scan, run CenterPoint on it, and evaluate using mean average precision (mAP) for vehicles. From our experiments, LidarDM achieves comparable mAP score (56.4\%) compared to real LiDAR (59.7\%), indicating strong semantic correlation of LidarDM's point cloud to ground-truth. We compute the mAP agreement between our generated LiDAR scan and raw LiDAR scan to be up to 81.1\%, showcasing a strong agreement between the two and demonstrating our approach's map-awareness and realism.

\subsubsection{Qualitative results}
We show qualitative results of our map-conditioned LiDAR sequence generation in Fig. \ref{fig:waymo_cond_qual}. Our generated results closely match the map conditioning, and the accumulated points over 90 frames highlights LidarDM's temporal consistency and map-awareness. The use of physical-based LiDAR sensor simulation guarantees that the generated point clouds are properly occluded by obstacles and appear as a realistic LiDAR sweep pattern. %

\subsubsection{Intensity Evaluation}

To assess our intensity realism, we use Wasserstein distance (WD) to compare the intensity distribution of our generated point cloud with that of the real-world point cloud. As a baseline, we adopt the physics-based intensity model described in Sec. \ref{sec:intensity}. Our experiment shows that LidarDM's intensity distribution matches well with the gt (WD of 0.0197), which significantly outperforms the physics-based baseline (WD of 0.0813).

\subsection{Augmenting Real Data with LidarDM}\label{sim2real} LidarDM is the first LiDAR generative model capable of generating data conditioned on semantic layouts. This capability offers the potential to augment the training data for and improve 3D perception and planning models.%

\subsubsection{Perception Model Sim2Real}

We first use LidarDM to generate around 70k frames of simulation data based on the layout from Waymo Dataset \cite{waymoopen}. After that, we pre-train a LiDAR-based 3D object detection model, CenterPoint \cite{centerpoint} (with PointPillars \cite{pointpillars} as its backbone), on these generated LiDAR frames, paired with the object labels from the dataset. We then train the same model on 35k frames of real data, both with and without the pre-training stage on the simulation data, to test the benefits of the LidarDM-generated data. LidarDM-augmented model achieves mAP score of 61.3\%, compared to 58.2\% achieved with the real data-only model. This shows that LidarDM is an effective generative data augmentation strategy, offering over 3\% improvement in detection accuracy.

\subsubsection{Planning Model Sim2Real}

Inspired by NMP \cite{nmp}, we developed a learning-based motion planner that takes the five most recent LiDAR observations (covering 0.5 seconds of past history) as input
and generates 10 frame (with 0.3 second intervals between each) cost map of the car's motion plan.
We make two changes to the original NMP model: 1) our planner does not require privileged HD Maps as input, allowing the experimental results to focus on the quality of our generated LiDAR, and 2) the planner does not explicitly incorporate the ego car's past trajectory \cite{codevilla, Dauner, zhairethinking}.
We employ a soft cross-entropy loss to train the cost map and sample from a trajectory bank (generated from the Waymo dataset using K-Means) during inference.

To show LidarDM's benefit for motion planning, we first train a model on 92k LidarDM-generated snippets, then fine-tune it on 9.2k real sequences. Expert driver trajectories serve as ground truth (GT) with traffic layout conditions generating LidarDM samples. For comparison, the same model is trained on only the 9.2k real sequences. Both models are trained for 30 epochs until convergence.

We report our results in Table \ref{tab:nmp_augmentation} 
Using generative pre-training improves the performance of the planner in a low-data regime. In particular, our collision rate after 3 seconds has been reduced by $32\%$ (relative).
To our knowledge, this is the first time conditional LiDAR generation has been shown to improve an end-to-end motion planner.

\begin{table}
\centering
\vspace{0.65em}

    \setlength{\tabcolsep}{4pt}
    \small\centering
    \begin{tabular}{lcccc}
        \toprule    
            Config & \makecell{L2 (m) \\  @ 1.0s} & \makecell{L2 (m) \\ @ 2.0s} & \makecell{L2 (m) \\ @ 3.0s} & \makecell{Collision \\ Rate (\%)}\\
        \midrule
            9.2k Real & \cellcolor{gold} $0.489$ & $1.374$ & $3.279$ & $1.65\%$\\
              \makecell[l]{9.2k Real+92k LidarDM} & $0.490$ & \cellcolor{gold} $1.341$ & \cellcolor{gold} $3.160$ & \cellcolor{gold} $1.12\%$\\
        \bottomrule
    \end{tabular}
\caption{
{\bf Planner data augmentation}: LidarDM-generated data can enhance performance of end-to-end planner. ($\fboxsep=0pt\fbox{\color{gold}\rule{2.5mm}{2.5mm}}$ indicates best)}
\label{tab:nmp_augmentation}
\vspace{-2em}
\end{table}

\section{Conclusion}

We presented \method, a novel layout-conditioned latent diffusion model for generating realistic LiDAR point clouds. Our approach frames the problem as a joint 4D world creation and sensory data generation task and develops a novel latent diffusion model to create 3D scenes. The resulting point cloud videos are realistic, coherent, and layout-aware.

\bibliographystyle{IEEEtran}

\bibliography{main}

\end{document}